\documentclass{article}

 \PassOptionsToPackage{numbers, compress}{natbib}

\usepackage[final]{neurips_2024}
\usepackage{graphicx}
\usepackage{subcaption}
\usepackage{wrapfig}
\usepackage{multirow}
\usepackage{longtable}

\usepackage{hyperref}
\usepackage{url}

\usepackage[utf8]{inputenc} % allow utf-8 input
\usepackage[T1]{fontenc}    % use 8-bit T1 fonts
\usepackage{hyperref}       % hyperlinks
\usepackage{url}            % simple URL typesetting
\usepackage{booktabs}       % professional-quality tables
\usepackage{amsfonts}       % blackboard math symbols
\usepackage{nicefrac}       % compact symbols for 1/2, etc.
\usepackage{microtype}      % microtypography
\usepackage{xcolor}         % colors

\title{DN-CL: Deep Symbolic Regression against Noise via Contrastive Learning}

\author{Jingyi Liu$^{12\dag}$\thanks{This work was supported in part by the National Natural Science Foundation of China under Grant 92370117, in part by CAS Project for Young Scientists in Basic Research under Grant YSBR-090.}~~,
Yanjie Li$^{12}$\thanks{Equal Contribution.}~~,
Lina Yu$^{1}$,
Min Wu$^{1}$,
Weijun Li $^{12}$\thanks{Corresponding Author.}~~,
Wenqiang Li$^{12}$,
Meilan Hao$^{1}$,\\
\textbf{
Yusong Deng$^{12}$,
Shu Wei$^{12}$}
\\
AnnLab, Institute of Semiconductors, Chinese Academy of Sciences\\
Haidian, Beijing, 100083, CN\\
University of Chinese Academy of Sciences\\
Huairou, Beijing, 101408, CN\\
\texttt{liujingyi@semi.ac.cn}
}

\begin{document}

\maketitle

\begin{abstract}
  Noise ubiquitously exists in signals due to numerous factors including physical, electronic, and environmental effects. Traditional methods of symbolic regression, such as genetic programming or deep learning models, aim to find the most fitting expressions for these signals. However, these methods often overlook the noise present in real-world data, leading to reduced fitting accuracy. To tackle this issue, we propose \textit{\textbf{D}eep Symbolic Regression against \textbf{N}oise via \textbf{C}ontrastive \textbf{L}earning (DN-CL)}. DN-CL employs two parameter-sharing encoders to embed data points from various data transformations into feature shields against noise. This model treats noisy data and clean data as different views of the ground-truth mathematical expressions. Distances between these features are minimized, utilizing contrastive learning to distinguish between 'positive' noise-corrected pairs and 'negative' contrasting pairs. Our experiments indicate that DN-CL demonstrates superior performance in handling both noisy and clean data, presenting a promising method of symbolic regression.
\end{abstract}

\section{Introduction}
\label{introduction}
Noise ubiquitously exists in signals due to physical, electronic, and environmental effects or measuring errors in experiment devices or equipment. Important discoveries often appear as simple, elegant, and concise expressions \cite{fong2023rethinking} which are refined from these observed signals, such as Johannes Kepler's discovery that Mars' orbit was an ellipse through the data tables on planetary orbits in 1601 \cite{udrescu2020ai}. \textit{Symbolic Regression (SR)} is a useful tool for finding the best expression to fit the data. Formally, given a set of data points (signals) $D=\{X,y\}\in \mathbb{R}^{n\times (d+1)}$, where $X\in \mathbb{R}^{n\times d}$ is the independent variable while $y\in \mathbb{R}^n$ indicate the dependent variable, SR aims to find a function that $y=f(X)$.

The current state-of-the-art methods are moved from the traditional \textit{Genetic Programming-based algorithms} \cite{mcconaghy2011ffx}\cite{2019Linear} to \textit{Deep Learning-based models} \cite{DBLP:conf/icml/BiggioBNLP21}\cite{10327762}. The deep symbolic regression methods are reported with superior results against the opponents. From the search strategy, we can divide the deep symbolic regression into two types: the end-to-end models and the large-scale pre-trained models. 

The end-to-end model searches the mathematical expressions in the search space from scratch for each SR task. For example, EQL \cite{kim2020integration} rectifies the activation functions of the fully connected neural networks and uses Lasso to control the connections. DSR \cite{petersen2021deep} employs an RNN to emit the probability distributions of the pre-order traversal, then sample skeletons, optimize constants, compute reward, and finally update the RNN with the risk-seeking policy gradients. Furthermore, DSO \cite{DBLP:journals/corr/abs-2111-00053} uses the neural network to guide the search for the Genetic Programming which achieves state-of-the-art results in SRBench \cite{DBLP:journals/corr/abs-2107-14351}.

However, the time cost of the end-to-end methods is huge. Some researchers utilize the large-scale model to learn the mapping relationship between the data points and mathematical expressions, such as SymbolicGPT \cite{DBLP:journals/corr/abs-2106-14131} and NeSymReS \cite{DBLP:conf/icml/BiggioBNLP21}. Once the model finished training, it can immediately output the predicted mathematical expression for a new problem. They achieved similar fitting accuracy with the end-to-end models but with low time cost in inference. Besides the advantage, the large-scale pre-trained models suffer from poor performances in out-of-training problems for they cannot rectify the wrongly predicted skeletons. Thus, some hybrid methods are proposed to enhance the prediction flexibility through combining reinforcement learning \cite{holt2023deep}\cite{liu2023snr}, Monte Carlo Tree Search \cite{DBLP:journals/corr/abs-2303-06833}, or the mixture of multiple methods \cite{DBLP:conf/nips/LandajuelaLYGSA22}\cite{xu2024reinforcement}.

Although real-world data abound with noise, the aforementioned methodologies are predicated on clean data, thus disregarding the nuances of practical scenarios. To train a noise-resistance large-scale pre-trained model, we think it requires a strong data encoder learning consistent data feature for different noise data that originated from same expression. Similar to the multi-view in computer vision task \cite{pmlr-v119-chen20j}, we treat the various noise data as different views of the ground truth expression. The different noisy data and clean data from the same ground truth expression are the positive samples, and the others are the negative samples. The positive samples should have similar data representations for they reflect the same ground truth expression. 

% Through our experiments, we observe that even when training the model with noisy data, the model also suffers from reduced fitting accuracy (see experiments in Section \ref{ablation_study}). We argue that this performance reduction comes from inconsistent learned feature representation of data. The different noise data that are derived from the same expression could have very distinct representations for there are no constraints in the model architecture to limit the feature representation. The inconsistent data representation makes the decoder learn the mapping relationship with a big challenge.

% We consider the critical point to be training a strong encoder that could output the consistent representation of data features for the noisy data that originated from the same expression. Similar to the multi-view in computer vision task \cite{pmlr-v119-chen20j}, we treat the various noise data as different views of the ground truth expression. The different noisy data and clean data from the same ground truth expression are the positive samples, and the others are the negative samples. The positive samples should have similar data representations for they reflect the same ground truth expression. 

To achieve noise resistant algorithm, we propose the model named \textit{\textbf{D}eep Symbolic Regression against \textbf{N}oise via \textbf{C}ontrastive \textbf{L}earning (DN-CL)}. DN-CL adopts two parameter-sharing encoders to embed the data points from different data transformations into data features. The two features are viewed as positive pairs and InfoNCE \cite{oord2018representation} are applied to reduce the distance between them. The decoder receives the data feature along with the pre-order traversal as the label to predict the mathematical expressions. The Cross-Entropy Loss is applied to guide the predictions. We learn the data feature and the pre-order traversal of the expressions in an end-to-end strategy rather than train a encoder first, then fix the parameter of encoder and update the decoder. Because the end-to-end strategy could learn better distinct features for SR task \cite{li2024mmsr}. In summary, our conclusion can be described as follows:
\begin{itemize}
    \item We propose to view the noise data and clean data as different views of the ground-truth mathematical expressions.
    \item We utilize contrastive learning to reduce the distance between features from clean data and noise data of the same expression while enlarging the distance between the negative pairs.
    \item We conduct sufficient experiments to evaluate the noise-resistant ability. Experiment results indicate that DN-CL shows better performance in both noisy and clean data.
\end{itemize}

\section{Related Works}
\label{related_works}
For simplicity, we term the algorithms that use deep-learning techniques to solve Symbolic Regression tasks as Deep Symbolic Regression. DN-CL contains two important components which are the pre-training SR model framework and contrastive learning. Therefore, this section introduces the related works of deep symbolic regression and contrastive learning.

\subsection{Deep Symbolic Regression}
\subsubsection{End-to-end models}
As the pioneer, EQL \cite{kim2020integration} substitutes the activation function of a neural network with primary operations such as $\sin$, $\cos$, etc. resulting in a large mathematical space that could degrade into specific expressions through a weight matrix. EQL employs Lasso \cite{tibshirani1996regression} to obtain matrix sparsity to achieve the simplicity of learned expressions. However, EQL faces challenges when addressing the expressions that contain division operations. The division operation brings numerical overflow if the dividend number is close to zero, leading to the gradient explosion.

DSR \cite{petersen2021deep} adopts risk-seeking policy gradients to update the network thus avoiding the gradient explosion problem introduced by division operation. It uses an RNN to emit the probability distribution of the pre-order traversal of the expression's binary tree. Then the expressions are sampled through the distribution. The constants are replaced by a placeholder $C$ and then to be optimized by BFGS \cite{fletcher2000practical} or rmsprop \cite{tieleman2012lecture}. DSR shows better results than its competitors on several public benchmarks. Further, to fully leverage the advantages of Genetic Programming \cite{mcconaghy2011ffx}\cite{2019Linear}-based algorithms and DSR, DSO \cite{DBLP:journals/corr/abs-2111-00053} proposes to use the neural network to guide the search of GP-based methods, which achieves the top-1 performance in SRBench \cite{DBLP:journals/corr/abs-2107-14351}.

In its experiments, DSR shows the ability to against noise because its reward function is the scaled normalized mean square error (NRMSE) that intrinsically forces the model to balance among uniform noise distribution. However, the search time cost is huge. Each iteration of DSR samples multiple expressions and optimizes the constants which are time-costly.

Another noticeable branch is AI Feynman \cite{udrescu2020ai}\cite{udrescu2020ai2} that fits a neural network first, then divides the origin problem into smaller pieces by prior knowledge of physics such as symmetry, separability, etc. They could rediscover the whole 100 basic physics equations. However, the strong prior knowledge results in poor performances in other benchmarks. Plus, the neural network may overfit the noise thus affecting the divide-and-conquer strategy while addressing the noisy data. 

\subsubsection{The large-scale pre-trained models}
\label{large_scale_pre-trained_model}
The large-scale pre-trained models are designed for accurate and fast inference of expression. They usually use an encoder to learn the data feature and employ a decoder to recursively output the expression. The training data are sampled from randomly generated mathematical expressions. The model could predict the expression with a high speed after the training is finished for it only needs a forward pass.

SymbolicGPT \cite{DBLP:journals/corr/abs-2106-14131} utilizes the text of the mathematical expression as the label. For example, the label of expression $2\cos(x)+y$ is $['2','c','o','s','(',')','+', 'y']$. The string-level encoding requires a larger amount of data to learn the semantics and syntax constraints of valid expressions. Therefore, an efficient representation would be using the pre-order traversal of the expression's binary tree. NeSymReS \cite{DBLP:conf/icml/BiggioBNLP21} implements such encoding strategy thus obtaining better performances. Furthermore, DeepSymNet \cite{10327762} designs a compact encoding of expression and views the SR as a classification task to achieve comparable results.

The above methods handled the problem of predicting the skeleton, but another main challenge is determining the constant value. The skeleton $C\cos(x)+Cy$, where $C$ is the constant placeholder, is attributed to different expressions with different values of $C$, leading to the ill-posed problem that different data distributions share the same label. Therefore, SymFormer \cite{vastl2022symformer} and E2E \cite{kamienny2022end} design the constant encoding methods to learn the constant in the training stage. The constant could be further optimized after decoding. They showed superior performances for the encoding of constants. For more personalized requirements, NSRwH \cite{DBLP:conf/icml/BendinelliBK23} trains a model outputting the expression that satisfies asked conditions, such as symmetry.

The large-scale pre-trained models are trained with the generated clean data, which ignores the reality that it is inevitable to tackle with noisy data. The model would suffer from decreased performances when handling noise data for good initial guess of expressions.

\subsubsection{Hybrid models}
The large-scale pre-trained models achieve fast inference speed while withstanding the low generalization. The performances are determined by the training set, the models would perform worse in out-of-training datasets. Furthermore, the prediction of the training data would not be accurate for all training expressions. Therefore, SNR \cite{liu2023snr} and GDSR \cite{holt2023deep} utilize the risk-seeking gradients to update the encoder to amend the skeleton. Another method TPSR \cite{shojaee2024transformer} incorporates the Monte Carlo Tree Search (MCTS) and the pre-trained model to obtain the generalization. Specifically, the neural network is used to guide the simulation process of MCTS.

With the wide generalization, the hybrid models have the same problem as the pre-trained large-scale models as illustrated in section \ref{large_scale_pre-trained_model}. They lack the components to tackle the noise data.

\subsection{Contrastive Learning for Symbolic Regression}
Contrastive Learning (CL) has obtained successful applications in computer vision, famous models include SimCLR \cite{chen2020simple}, MoCo \cite{he2020momentum}, ALBEF \cite{li2021align} etc. Considering the achievements of CL, some researchers use CL to address specific problems in SR.

T-JST \cite{li2022transformer} proposes to utilize the CL to distinguish the features of different expressions that share the same skeletons. It is based on the intuition that the data features of expressions that originate from the same skeleton should be similar, and the data features of expressions that originate from different skeletons should be dissimilar. It alleviates the ill-posed problem. However, the ill-posed problem is solved with the proposal of constant encoding \cite{vastl2022symformer}\cite{kamienny2022end}.

SNIP \cite{meidani2023snip} utilizes the multimodal algorithm to process SR tasks. It uses a two-step training strategy, the first step uses CL to learn the data feature to distinguish each expression, then train a decoder to predict expressions. The experiments indicate the learned data feature by CL contributes to accurate prediction. Furthermore, MMSR \cite{li2024mmsr} views the input and output of SR tasks as different modalities. It trains a model in an end-to-end way and uses the CL to align the data feature and label feature learned by encoders. A decoder is applied to receive the two features to predict expressions. The experiments show the alignment of different modalities could bring significant improvements. 

However, such as the pre-trained models above, they do not take into account the problem of noisy data. We propose a framework for using contrastive learning to address the noise data of the large pre-trained models. The framework could be implemented in any pre-trained model as long as it has an encoder and decoder.

\section{Methods}
\label{methods}
To obtain a noise-resistant large-scale pre-trained model, our model involves a typical pre-trained model for symbolic regression modified with a noise process component to address noise data. We first describe the structure DN-CL, then introduce the implementation details.

\subsection{DN-CL}

\begin{figure}
    \centering
    \includegraphics[width=\linewidth]{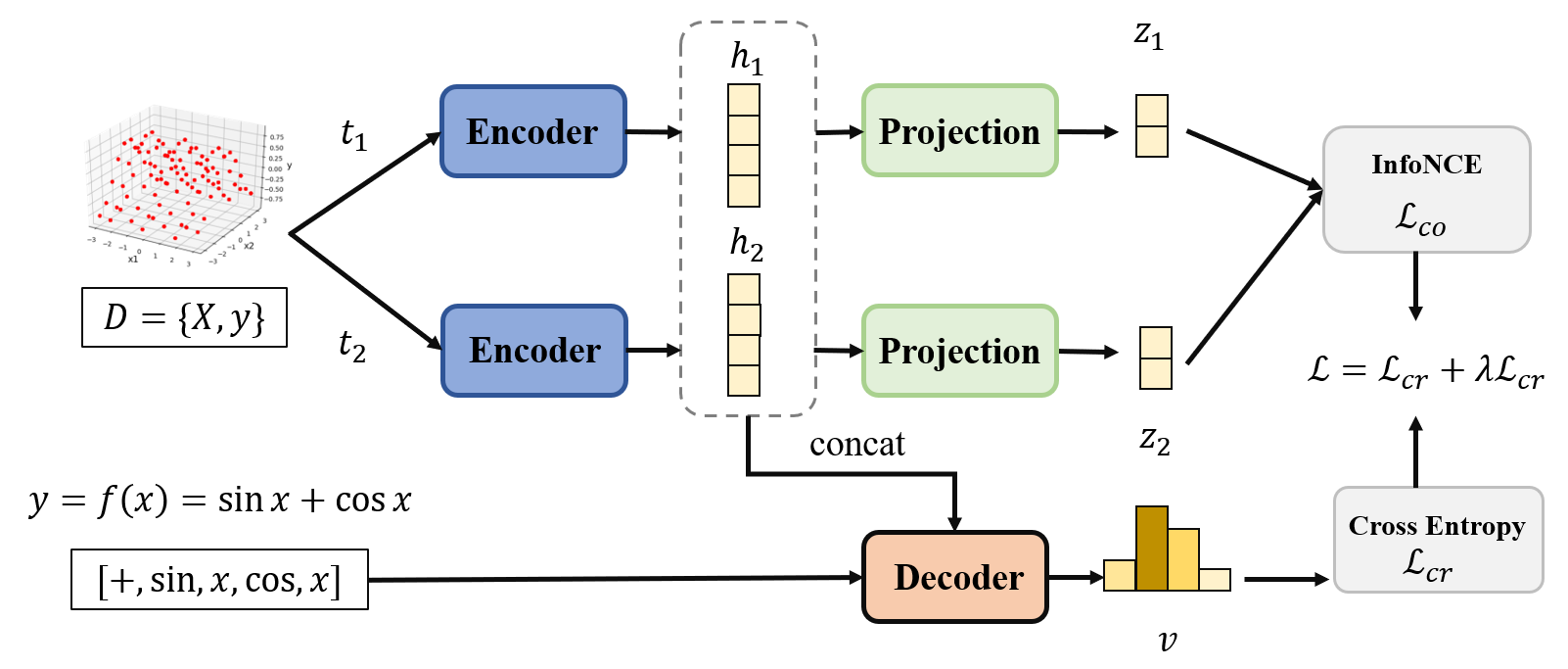}
    \caption{The framework of DN-CL. The input $D$ is fed to two parameter-sharing encoders with two different transformations. $t_1$ denotes identity operation meaning no transformation is made. $t_2$ denotes adding noise into the input $D$ to get a different view of true mathematical expression. The hidden features $h_1$ and $h_2$ are the input of a projection layer obtaining vectors $z_1$ and $z_2$ to compute InfoNCE loss. Besides, we concat $h_1$ and $h_2$ as the data feature along with the expression label inputting to the decoder to predict the expression. The loss is composed of the InfoNCE and cross-entropy loss.}
    \label{fig:framework}
\end{figure}

The structure of DN-CL is drawn in Figure \ref{fig:framework}. Two different operations $t_1$, $t_2$ are implemented on the clean input data $D=\{X,y\}\in \mathbb{R}^{n\times (d+1)}$ to obtain noise data, where $n$ is the number of data points and $d$ is the number of variables. $t_1$ and $t_2$ are viewed as different views of the ground truth expression. Specifically, we set $t_1$ as an identity operation which means no operation is conducted on clean data $D$. In contrast, we add noise in dependent value $y$ through transformation $t_2$ as follows:
\begin{equation}
y' = y+{\rm Gaussian}(\mu, \sigma^2)=y+\frac{1}{\sigma\sqrt{2\pi}}e^{-\frac{(y-\mu)^2}{2\sigma^2}}
\end{equation}

In this instance, Gaussian noise with a mean $\mu=0$, and a standard deviation $\sigma$ that scales with the root-mean-square of the dependent variable, is added:
\begin{equation}
    \sigma = \eta \sqrt{\frac{1}{n}\sum_{i=1}^{n}y_i^2}
\end{equation}
where $\eta$ controls the noise level. We then feed the two transformed data into two parameter-sharing encoders $f_{\rm Encoder}$ obtaining two data features $h_1=f_{\rm Encoder}(t_1(D))$ and $h_2=f_{\rm Encoder}(t_2(D))$. If $h_1$ and $h_2$ are origins from the same expression, then they are positive pairs, otherwise they are negative pairs. The positive pairs should be similar to each other while the negative pair should be dissimilar to each other. Therefore, we utilize InfoNCE \cite{oord2018representation} to reduce the distance between positive pairs while increasing the distance between the negative pairs:
\begin{equation}
    \mathcal{L}_{\rm co}=-\log\frac{\exp{(qk^{+}/\Gamma)}}{\exp{(qk^+/\Gamma)}+\sum_{k^-}\exp{(qk^-/\Gamma)}}
\end{equation}
where $q=f_{\rm proj}(h_1)$ is the ``query'', the goal of learning is to retrieve the corresponding ``key''  $k=f_{\rm proj}(h_2)$. We add a projection layer $f_{\rm proj}$ to reduce the dimension of $h_1$ and $h_2$ for empirical results showing it is an effective strategy for improving fitting performance in vision tasks \cite{chen2020simple}. The set $\{k^-\}$ consists of the negative samples while $\{k^+\}$ is the set of positive pairs of $q$. $\Gamma$ is a temperature hyper-parameter \cite{wu2018unsupervised} for $l_2$-normalized $q$, $k$.

Normally we use $h_1$ and $h_2$ to predict tasks as suggested in reference \cite{chen2020simple}. In this study, we concat $h_1$ and $h_2$ feeding to the decoder for two reasons. First, we expect the decoder could accurately predict the expressions through noise data, thus we put the noise feature $h_2$ into the decoder to increase the learning difficulty. Second, we also expect the model still maintain superior results on clean data, thus the $h_1$ are feed to the decoder too. In practical, we replace half of the batch with the data after transformation $t_2$ for coding simplicity.

We use the Cross-Entropy Loss to supervise the learning of expressions. Expressions are normally represented by the pre-order traversal of the binary tree. Assume the expression is represented by a traversal $\tau=[\tau_1,\tau_2,...,\tau_L]$, where $L$ is the maximum length of the pre-order traversal, we compute the average cross-entropy loss for all tokens:
\begin{equation}
    \mathcal{L}_{\rm cr}=-\frac{1}{L}\sum_{i=1}^L\sum_{c=1}^{C}l_{i,c}\log{p_{i,c}}
\end{equation}
where $C$ is the number of operators, $l_{i}$ is the true label of the token $\tau_i$, $p_{i}$ is the output of the decoder. Finally, the whole loss of DN-CL is the sum of InfoNCE loss $\mathcal{L}_{\rm co}$ and cross-entropy loss $\mathcal{L}_{\rm cr}$:
\begin{equation}
    \mathcal{L} = \mathcal{L}_{\rm cr}+\lambda\mathcal{L}_{\rm co}
\end{equation}
where $\lambda$ is the hyperparameter to control the effect of contrastive loss $\mathcal{L}_{\rm co}$.

\subsection{Implementation Details}
\textbf{Encoder and Decoder}. Generally, we propose a framework that could alleviate the perturbation taken by noise data. This framework has no specific requirement for encoder and decoder. The encoder and decoder could be replaced by any advanced structures such as NeSymReS \cite{DBLP:conf/icml/BiggioBNLP21}, E2E \cite{kamienny2022end}, etc. We adopt the SymFormer \cite{vastl2022symformer} for its efficient constant encoding strategy which solved the  ``ill-posed'' problem. The constants are encoded using a scientific-like notation where a constant $C\approx c_m10^{c_e}$ is represented as a tuple of the exponent $c_e$ and the mantissa $c_m$. In this representation, the mantissa is in the range $[-1,1]$, and the exponent is an integer. For example,
the expression $0.017x + 1781.5$ will have symbols $[+, \times, x, C-1, C4]$ and constants $[0, 0, 0, 0.17,0.17815]$. To further help the model represent constants, we add all integers from interval $[-5,5]$ into the model vocabulary following the origin work \cite{vastl2022symformer}.

\textbf{Gold standard.} When evaluating the obtained mathematical expressions on noise data by the $R^2$, it is hard to judge whether the results are overfitted to the noise or obtaining high accuracy for finding the ground truth expressions. It is assumed that the ground truth expression could exhibit the stable and outstanding performances across different sampled noise data \cite{fong2023rethinking}. Therefore, We set a gold standard by applying the ground truth expressions to the data with different noise levels to evaluate the model performance. If the model has similar performance on various noise level data with the ground truth, we say the model is better.

\section{Experiments}
\label{experiments}

\subsection{Dataset}
\textbf{Training data}. We follow the expression generation process with Symformer \cite{vastl2022symformer}. Specifically, we use two independent models to predict the uni-variate and bi-variate tasks, respectively. For the uni-variate training set, we generate 5,000,000 expressions and each expression contains 500 points while training 100 points are sampled from the data points randomly in each iteration. For the bi-variate, we generate 10,000,000 expressions for the additional variable, which increases the search space and thus needs more data to train. 500 points are sampled for each expression and the training process uses randomly sampled 200 points. We use an extra 100,000 expressions as the validation sets for each model. Then, the noise is added by different levels of noise. In each batch, the noise is sampled in the output dimension through $y'=y+{\rm Gaussian}(\mu,\sigma^2)$, where $\mu=0,\sigma=\eta{\rm RMS}(y)$ with $\eta=0.1$.

\textbf{Evaluation data}. We evaluate our model on public benchmarks collected from Nguyen \cite{uy2011semantically}, R \cite{krawiec2013approximating}, Livermore \cite{petersen2021deep}, Koza \cite{koza1994genetic}, Keijzer \cite{keijzer2003improving}. To test the model performances on noisy data, the hyperparameter $\eta$ is set by different numbers $\eta=0.0,0.001,0.01,0.1$ to evaluate the noise endurance abilities. The physical dataset AI Feynman \cite{udrescu2020ai} is included by selecting the equations whose variable number less or equal to 2. Detailed benchmark information is in Appendix \ref{appendix:bechmark_info}.

\subsection{Setting}
\subsubsection{Metric}
\textbf{Fitting accuracy.} We use the Coefficient of Determination ($R^2$) to assess the quality of fitting results. The $R^2$ is expressed as follows:
\begin{equation}
    R^2(y,\hat{y})=1-\frac{\sum_{i=1}^n(y_i-\hat{y}_i)^2}{\sum_{i=1}^n(y_i-\bar{y})^2},
\end{equation}
where $y_i$ and $\hat{y_i}$ are the ground-truth and predicted values for point $i$, respectively. $\bar{y}$ is the average of $y_i$ over all points. $n$ is the number of test points. The advantage of using $R^2$ is its simple interpretation. $R^2>0$, implies that the predicted value is better than the average value; $R^2=1$, implies that the algorithm obtained the perfect fit; and $R^2<0$ implies that the predicted value is worse than the average value. Each expression is conducted 10 independently run. Maximum $R^2$ and mean $R^2$ were calculated to evaluate the algorithm performances.

\textbf{Expression complexity.} For the sake of simplicity, we follow the work of SRBench \cite{DBLP:journals/corr/abs-2107-14351} which defines complexity as the number of mathematical operators, features and constants in the model, where the mathematical operators are in the set $\{+,-,\times,\div,\sin,\cos,\arcsin,\arccos,\exp,\log,{\rm pow}\}$. In addition to calculating the complexity of the raw model forms returned by each method, we calculated the complexity of the models after simplifying via {\fontfamily{qcr}\selectfont
Sympy}\footnote{https://www.sympy.org/}.

\subsubsection{Hyperparameters}
For the training of our method, we set the parameter $\lambda$ as 0.1, and the noise level of training set is $\eta=0.1$. We implement the NeSymReS\footnote{https://github.com/SymposiumOrganization/NeuralSymbolicRegressionThatScales}, E2E\footnote{https://github.com/facebookresearch/symbolicregression.}, T-JST\footnote{https://github.com/AILWQ/Joint\_Supervised\_Learning\_for\_SR.} and DeepSymNet\footnote{https://github.com/wumin86/DeepSymNet} by their official implementations. For the end-to-end methods, we scratch their results provided by SRBench and compare our method with them. Details of the hyperparameters of our model and the baselines are in Appendix \ref{implement_details} and Appendix \ref{baseline_details}, respectively.

To evaluate the stability of the trained model with data permutation, we generate 10 independent data point sets for each public expression. The noise is set at different levels: [0.1, 0.001, 0.01, 0.0]. The maximum $R^2$ is recorded as the best result.

\subsection{Ablation Study}
\label{ablation_study}

\begin{wrapfigure}{r}{0.6\textwidth}
  \centering
  \includegraphics[width=0.58\textwidth]{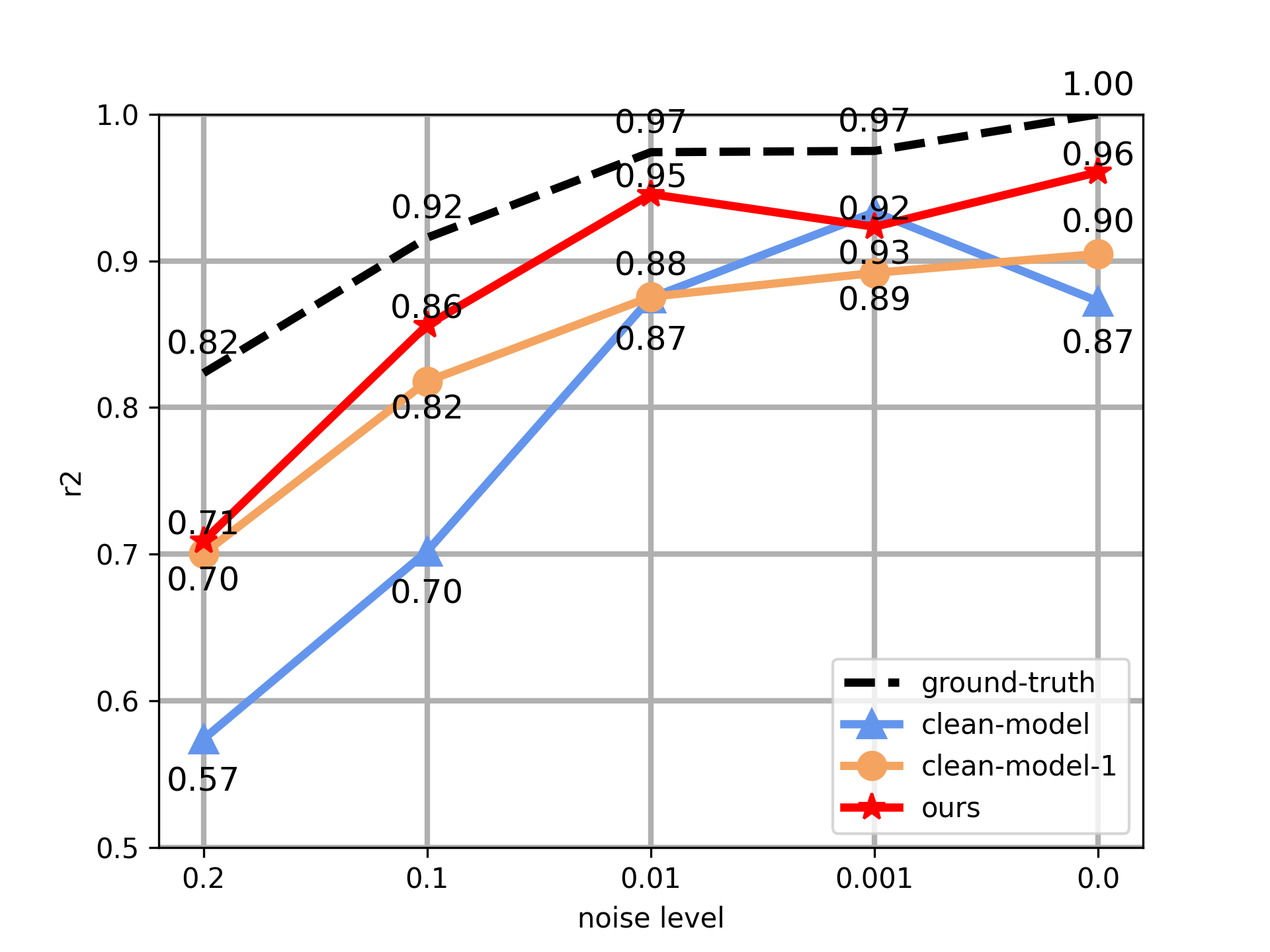}
  \caption{The average $R^2$ of each methods in uni-variate benchmarks. The x-axis indicates descending noise level from 0.2 to 0.0. The y-axis represents the average $R^2$ across all benchmarks. The dash line, blue, orange, and red line represents ground truth, clean-model, clean-model-1, and our models, respectively.}
  \label{avg_bivariate}
\end{wrapfigure}

\textbf{Effects of each component.} We conduct experiments to validate the effect of noise data and contrastive loss, respectively. The results of uni-variate benchmarks are in Figure \ref{avg_bivariate}.

The results of ``ground-truth'' are calculated by the $R^2$ in the noise data using the ground truth expressions. The ``clean-model'' indicates we train the model without noise data and contrastive loss. For detail comparison, we also train the ``clean-model-1'' by using the mixture of half noise data and clean data without contrastive loss. With the reduction of noise level, basically all methods achieved increased fitting accuracy. Clean models have some resistance to noise when the level of noise is low, such as $\eta=0.001$, but it suffers huge descend when the noise level increased even though the noise level is tolerable (e.g. $\eta=0.01$). Adding noise data into the training set provides valuable effect against noise, when the noise level increased to 0.1, ``clean-model-1'' exhibits significant improvement compared with ``clean-model''. Our model displayed more accurate prediction by using the contrastive loss to learn consistent data features. Note that, we only add the noise data with $\eta=0.1$, but the models trained with noise data unfold better $R^2$ in unseen data (e.g. noise level equals to 0.2). It indicated that adding noise data into training could enhance the model's scalability.

\begin{figure}[htbp]
    \centering
    \begin{subfigure}[b]{0.3\textwidth}
        \includegraphics[width=\textwidth]{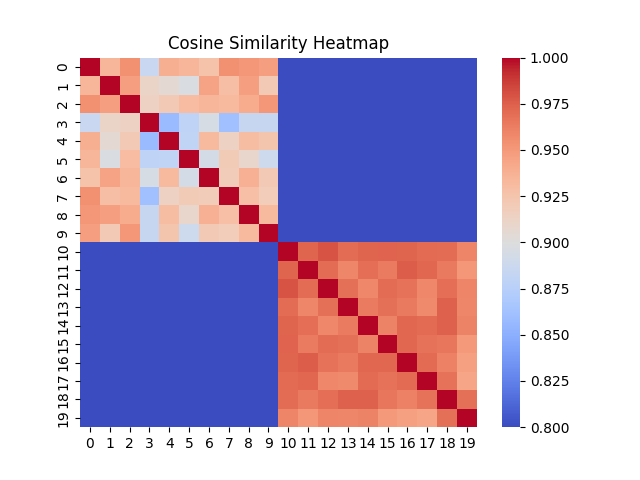}
        \caption{Clean model}
        \label{Keijzer-15:clean-model}
    \end{subfigure}
    \hfill
    \begin{subfigure}[b]{0.3\textwidth}
        \includegraphics[width=\textwidth]{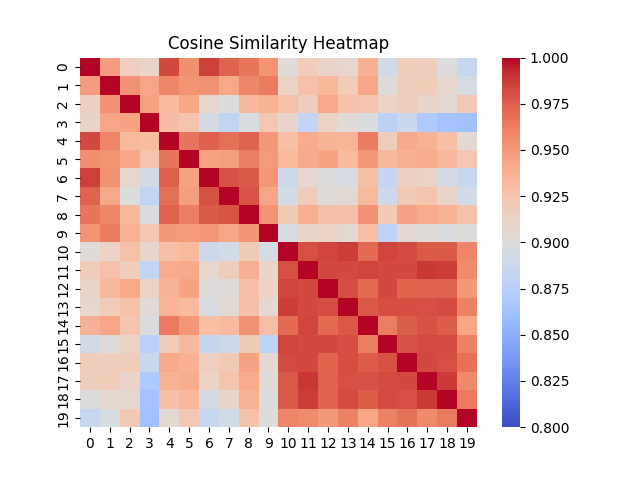}
        \caption{Clean-model-1}
        \label{Keijzer-15:clean-model-1}
    \end{subfigure}
    \hfill
    \begin{subfigure}[b]{0.3\textwidth}
        \includegraphics[width=\textwidth]{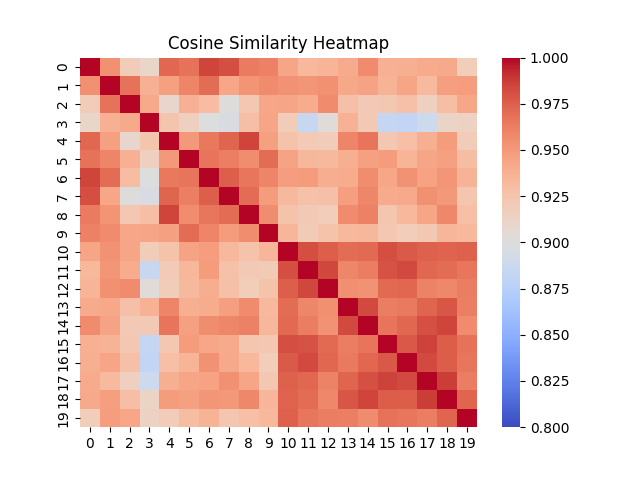}
        \caption{Ours}
        \label{Keijzer-15:our-model}
    \end{subfigure}

    \caption{Heatmaps for the features outputted by the encoder between clean data and noise data (noise level $\eta=0.1$). The expression is Keijzer-15.}
    \label{Heatmap:Keijzer-15}
\end{figure}

\textbf{Heatmaps of different noise level data features.} Our model achieves the highest $R^2$ against the components, which is the closest to the ground truth. We attribute the reason to the consistent features learned by contrastive loss with different noise data. The consistent features alleviate the work of decoder to accurate prediction. To illustrate this point, we display the heatmaps for the data features outputted by the encoder between the clean data and noise data with noise level 0.1. For the clean data and noise data, we sample 10 independent datasets with different random seeds, each dataset contains 100 data points. These datasets are fed into the encoders of clean model and our model, results in $h_{\rm noise}, h_{\rm clean}\in\mathbb{R}^{10\times32\times384}$. Then we sum the metrixes in the second dimension and contact them in the first dimension obtain $h\in\mathbb{R}^{20\times384}$. Finally, we use the cosine distance to calculate the similarities between these features, the results are shown in Figure \ref{Heatmap:Keijzer-15}.

The heatmaps show four distinct areas: the upper left and bottom right are the similarities between the noise datasets and the clean datasets, respectively; the upper right and bottom left are the similarities of between noise datasets and clean datasets. The clean model exhibits high similarity between clean datasets, poorer similarity between noise datasets and bad similarity between clean and noise datasets thus results in huge $R^2$ reduction when tested on high noise level data. However, the clean-model-1 shows better feature similarity between the noise data and clean data that leads to higher $R^2$. Furthermore, our model utilize the contrastive learning as a constraint to reduce the distance between the features of data with different noise level.

\begin{wraptable}{l}{0.5\textwidth}
    \centering
    \caption{Recovery rate on Nguyen1. 10 independent experiments are conducted. The data are generated by 10 random seeds.}
    \label{recovery_rate}
    \begin{tabular}{ccccc}
        \hline
        \multirow{2}{*}{\textbf{Methods}} & \multicolumn{4}{c}{\textbf{Recovery rate}} \\ \cline{2-5}
        & 0.1 & 0.01 & 0.001 & 0.0 \\\hline
        clean-model & 0\% & 0\% & 20\% & 80\% \\
        clean-model-1 & 0\% & 0\% & 20\% & 30\% \\
        ours & \textbf{30\%} & \textbf{50\%} & \textbf{20\%} & \textbf{80\%} \\ \hline
    \end{tabular}
\end{wraptable}

\textbf{Recovery rate for case study.} Recovery rate is a harder evaluation metric compared with the $R^2$ for it requires exactly recover the ground truth expression, it can reflect the model ability against noise. We calculate the recovery rate on ``Nguyen1'', results are shown in Table \ref{recovery_rate}. Ours model achieves highest recovery rate across different levels of noise. The clean model performed well with clean data and low-level noise, but could not recover the expression with higher noise level. The clean-model-1 even has the worst recovery rate in clean data that indicates the noise data affect the training without the contrastive learning.

\subsection{Compare with the State-of-the-arts}

\begin{table}[]
    \centering
    % \footnotesize
    \caption{Compared with state-of-the-arts. The number ``0.0'' indicates experiment on clean data, whereas ``0.1'' means the results with noise level $\eta=0.1$.}
    \label{compared_with_pre_trained}
    \resizebox{\textwidth}{!}{%
    \begin{tabular}{ccc cc cc cc cc}
    \hline
    \multirow{2}{*}{\textbf{Dataset}} & \multicolumn{2}{c}{\textbf{Ours}} & \multicolumn{2}{c}{\textbf{NeSymReS}} & \multicolumn{2}{c}{\textbf{E2E}} & \multicolumn{2}{c}{\textbf{T-JST}} & \multicolumn{2}{c}{\textbf{DeepSymNet}} \\ \cline{2-11}
     & \textbf{0.0} & \textbf{0.1} & \textbf{0.0} & \textbf{0.1} & \textbf{0.0} & \textbf{0.1} & \textbf{0.0} & \textbf{0.1} & \textbf{0.0} & \textbf{0.1}\\
    \hline
    \textbf{Keijzer} &  0.9880 &  0.9674 $\downarrow$ &  0.8734 &  0.9406 $\uparrow$ &  0.8753 &  0.7048 $\downarrow$  & 0.8497  & 0.8934 $\uparrow$  & 0.8117   &  0.7647 $\downarrow$ \\
    \textbf{Koza}     &  1.0000 &  0.9903 $\downarrow$ &  1.0000 &  0.6873 $\downarrow$ &  1.0000 &  0.9733 $\downarrow$  & 0.9449 & 0.9232 $\downarrow$    & 0.8504   & 0.8939 $\uparrow$  \\
    \textbf{Nguyen} &  0.9997 &  0.9135 $\downarrow$ &  0.9259  &  0.8937 $\downarrow$  &  0.6961 &  0.7411 $\uparrow$  & 0.8632 & 0.8777 $\uparrow$    & 0.7966   &  0.8358 $\uparrow$ \\
    \textbf{Constant}  &  0.8749 &  0.7749 $\downarrow$ &  0.9064 &  0.8354 $\downarrow$ &  0.7265 &  0.7561 $\uparrow$ & 0.7785  & 0.8025 $\uparrow$    &  0.7379  & 0.7241 $\downarrow$   \\
    \textbf{Livermore} &  0.9396 &  0.8050 $\downarrow$ &  0.8300 &  0.7962 $\downarrow$  &  0.7544  &  0.6661 $\downarrow$ & 0.9131 & 0.9084 $\downarrow$    & 0.7611 & 0.6864 $\downarrow$ \\
    \textbf{R}         &  0.9990 &  0.9885 $\downarrow$ &  0.6665 &  0.9840 $\uparrow$ &  0.9988 &  0.9690 $\downarrow$ & 0.9944 & 0.9880 $\downarrow$    &  0.9963  & 0.9893 $\downarrow$  \\ \hline
    \textbf{Avg}       & \textbf{ 0.9670} &  \textbf{0.9066 $\downarrow$} &  0.8670 &  0.8562 $\downarrow$ &  0.8419 &  0.8017 $\downarrow$ & 0.8906 & 0.8989 $\uparrow$ & 0.8257 &   0.8157 $\downarrow$\\
    \hline
    \end{tabular}%
    }
\end{table}

% \begin{wrapfigure}{l}{0.6\textwidth}
%   \centering
%   \includegraphics[width=0.58\textwidth]{./figs/Livermore.png}
%   \caption{The average $R^2$ of each methods for Livermore dataset. The x-axis indicates descending noise level from 0.2 to 0.0. The y-axis represents the average $R^2$ across all benchmarks.}
%   \label{livermore}
% \end{wrapfigure}

We compare our model with several advanced pre-trained models including NeSymRes\cite{DBLP:conf/icml/BiggioBNLP21}, E2E\cite{kamienny2022end}, T-JST\cite{li2022transformer}, DeepSymNet\cite{10327762} on public datasets. The $R^2$ on clean data (with $\eta=0.0$) and noise data (with $\eta=0.1$) are shown in Table \ref{compared_with_pre_trained}.

\begin{figure}[htbp]
    \centering
    \begin{subfigure}[b]{0.45\textwidth}
        \includegraphics[width=\textwidth]{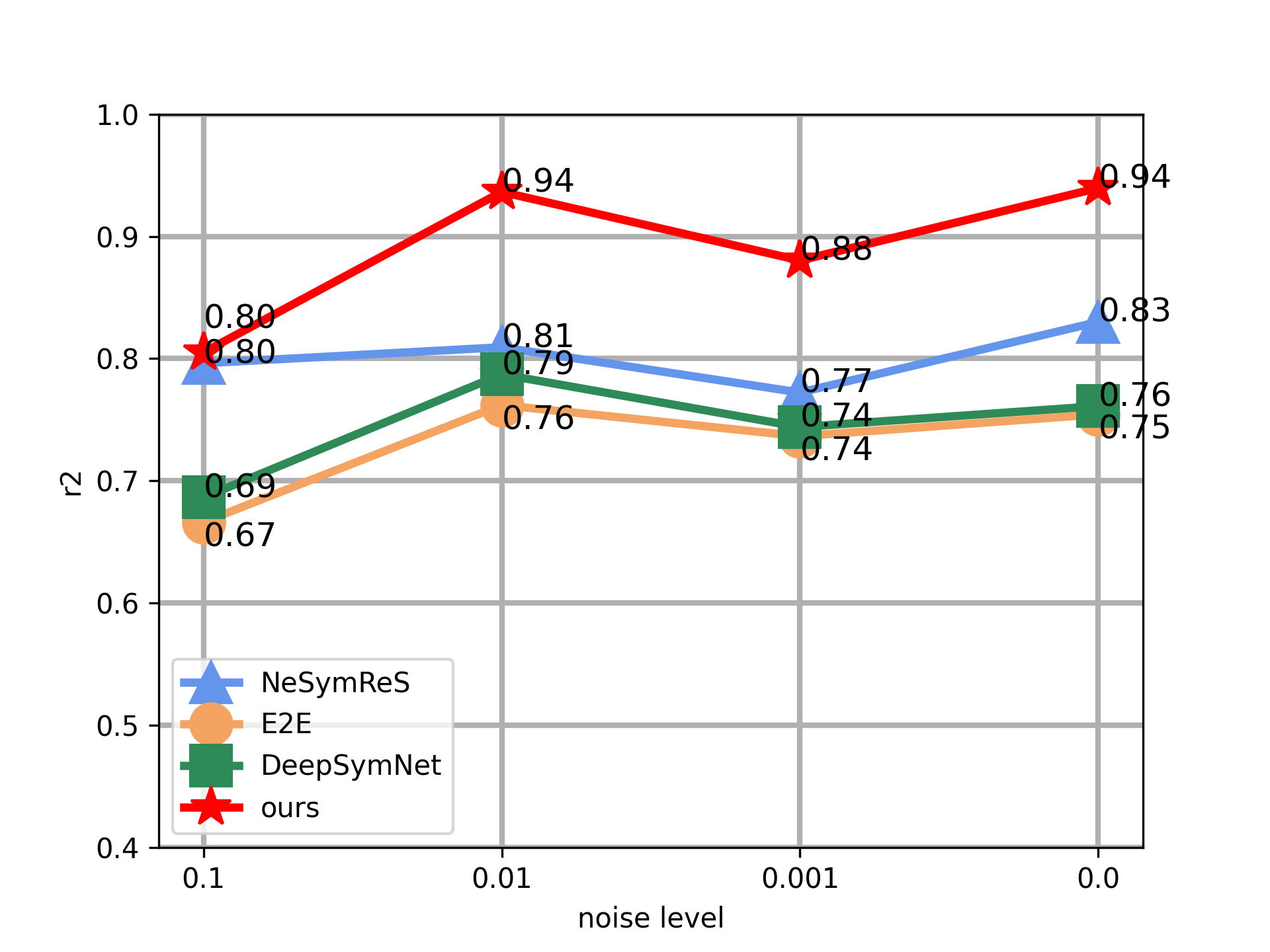}
        \caption{Livermore}
        \label{Livermore_main}
    \end{subfigure}
    \hfill
    \begin{subfigure}[b]{0.45\textwidth}
        \includegraphics[width=\textwidth]{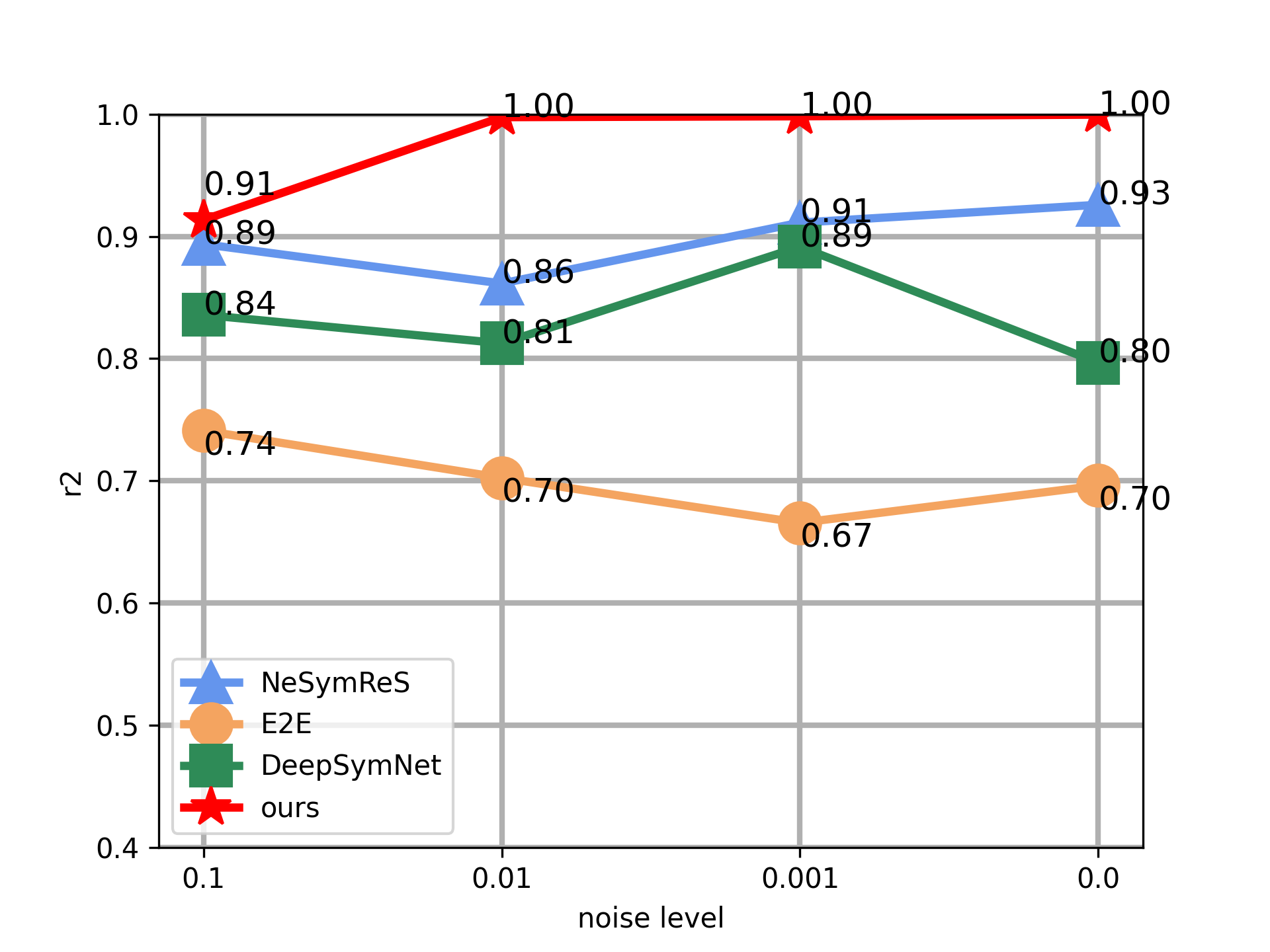}
        \caption{Nguyen}
        \label{Nguyen_main}
    \end{subfigure}

    \caption{The average $R^2$ of each methods for Livermore and Nguyen datasets. The $x$-axis indicates descending noise level from 0.1 to 0.0. The $y$-axis represents the average $R^2$ across all benchmarks.}
    \label{livermore_and_nguyen_and_keijizer}
\end{figure}

\textbf{Statistics of fitting accuracy.} Overall, almost all methods suffer through fitting accuracy discount when processing with noise data. Because noisy data brings difficulties for predicting accurate skeletons. Plus, the complexity of the predicted symbolic expressions are limited, it is hard to fit all data points with noise data. For the average $R^2$, our model achieves the highest $R^2$ compared with others on both clean data and noise data. It indicates that the introducing of noise data not affects the training of clean data. We contribute this to our training strategy that use half clean data and half noise data to train the model.

\textbf{Performance under different noise level data.} The pre-trained model are trained on training data with noise level $\eta=0.1$, while Table \ref{compared_with_pre_trained} indicates good performances in-domain data, we conduct experiments on out-of-domain data with noise level 0.001 and 0.01. Results are displayed in Figure \ref{livermore_and_nguyen_and_keijizer}. With the noise level reduced, all the methods displayed ascending lines. Within these methods, our model achieves highest $R^2$. Details of other datasets are in Appendix \ref{Complementary_experimental_results}.

\begin{figure}
    \centering
    \includegraphics[width=0.9\textwidth]{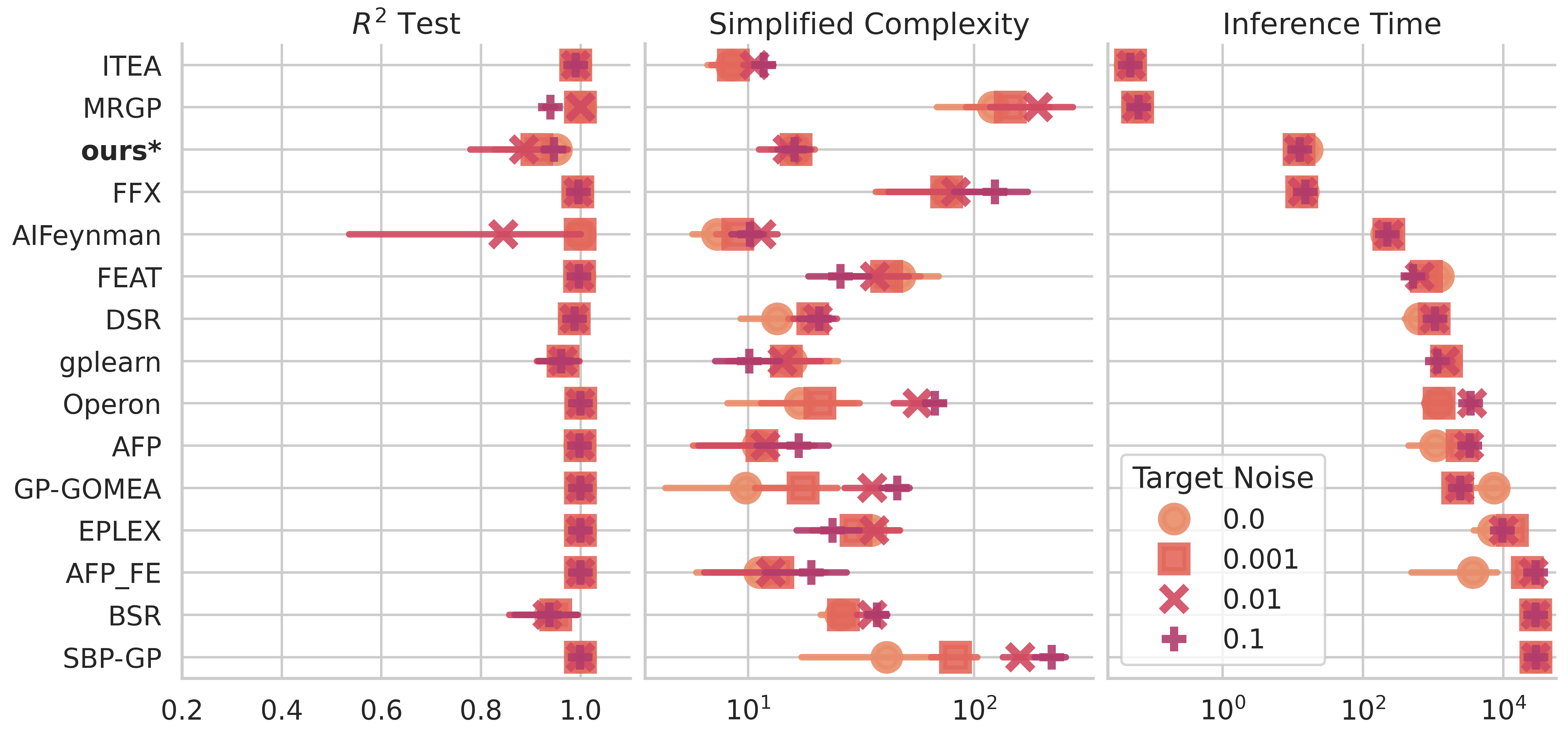}
    \caption{Results on AI Feynman datasets, $R^2$, simplified complexity, and inference time are displayed. Color/shape indicates level of noise added to the target variable.}
    \label{srbench}
\end{figure}

\textbf{Comparison with end-to-end models.} Further, we compare our method with end-to-end models which are flexible in handing out-of-domain data. The $R^2$, simplified complexity, and inference time are shown in Figure \ref{srbench}. We sort the algorithms by the ascending order of inference time.

The end-to-end model posses superior out-of-domain fitting accuracy for their inherent nature. Therefore, they almost perform consistent good performances across various noise levels. However, they trade the fitting performance with higher complexities. Complex symbolic expressions opt to overfit the noise data thus not reflect the ground truth expressions. Plus, over complexity reduced the interpretability. Furthermore, the end-to-end models are time-consuming when handing new tasks, while the pre-trained model such as ours has superior advantages in inference time. Overall, our model achieves a good trade between fitting accuracy, expression complexity, and inference time.

\section{Conclusion and future works}
\label{conclusion}
\textbf{Conclusion.} In this paper, we propose to handle with noise data by large-scale pre-trained model for symbolic regression. Through viewing the noise data as different views of the ground truth expressions, the contrastive learning is adopted to lean consistent data features across various noise data. Specifically, we use two parameter-sharing encoders to map the data points into the data features, then InfoNCE loss is utilized to enlarge the distance between the data points derived from different expressions, while enclose the distance between the data points originated from the same expressions. Finally, the concated data features from two encoders are feed into a decoder to obtain expressions. Ablation study indicates the importance of noise data in training and the contrastive loss. Compared with other end-to-ed state-of-the-arts, our model achieves good balance by considering fitting accuracy, expression complexity, and inference time.

\textbf{Future works.} The focus of our model are training a strong encoder that can resist noise. While the current trade of deep symbolic regression is hybrid method, we could apply our encoder to the hybrid model that contains a data encoder to obtain more accurate results.


\bibliographystyle{plainnat}  
\bibliography{reference.bib}

\newpage

\appendix
\section{Details of benchmarks}
\label{appendix:bechmark_info}
\setcounter{table}{0} 
\renewcommand{\thetable}{A\arabic{table}}

The details of several public datasets are listed in Table \ref{dedails_of_public_datasets}. The expression name, mathematical form, and the number of variable are included. Plus, the details of used physical dataset AI Feynman are listed in Table \ref{AIFyenman}.

\begin{footnotesize}
\begin{longtable}{@{\extracolsep{\fill}}ccc@{}}
    \caption {Benchmark symbolic regression problem specifications. Input variables are denoted by $x$ and\/or $y$.}
    \label{dedails_of_public_datasets}\\\hline
    \textbf{Dataset} & \textbf{Expression} & \textbf{Variable number} \\\hline
    \endfirsthead
    \multicolumn{3}{c}{Table \ref{dedails_of_public_datasets} continued}\\\hline
    \endhead
    \hline
    \endfoot
    \hline
    \endlastfoot
    Nguyen1 & $x^3+x^2+x$ & 1  \\
    Nguyen2 & $x^4+x^3+x^2+x$ & 1 \\
    Nguyen3 & $x^5+x^4+x^3+x^2+x$ & 1 \\
    Nguyen4 & $x^6+x^5+x^4+x^3+x^2+x$ & 1\\
    Nguyen5 & $\sin(x^2)\cos(x)-1$ & 1 \\
    Nguyen6 & $\sin(x)+\sin(x+x^2)$ & 1 \\
    Nguyen7 & $\log(x+1)+\log(x^2+1)$ & 1 \\
    Nguyen8 & $\sqrt{x}$ & 1 \\
    Nguyen9 & $\sin(x)+\sin(y^2)$ & 2 \\
    Nguyen10 & $2\sin(x)\cos(y)$ & 2 \\
    Nguyen11 & $x^y$ & 2 \\
    Nguyen12 & $x^4-x^3+0.5y^2-y$ & 2  \\
    \hline
    Constant1 & $3.39x^3+2.12x^2+1.78x$ & 1 \\
    Constant2 & $\sin(x^2)\cos(x)-0.75$& 1 \\
    Constant3 & $\sin(1.5x)\cos(0.5y)$  & 2 \\
    Constant4 & $2.7x^y$ & 2 \\
    Constant5 & $\sqrt{1.23x}$ & 1 \\
    Constant6 & $x^{0.426}$ & 1 \\
    Constant7 & $2\sin{(1.3x)}\cos(y)$ & 2 \\
    Constant8 & $\log(x+1.4)+\log(x^2+1.3)$ & 1 \\
     \hline
    R1 & $\frac{(x+1)^3}{x^2-x+1}$ & 1 \\
    R2 & $\frac{x^5-3x^3+1}{x^2+1}$ & 1 \\
    R3 & $\frac{x^6+x^5}{x^4+x^3+x^2+x+1}$ & 1 \\
     \hline
    Livermore-4 & $\ln{(x+1)}+\ln{(x^2+1)}+\ln{(x)}$ & 1 \\
    Livermore-5 & $x^4-x^3+x^2-y$ & 2 \\
    Livermore-9 & $x^9+x^8+x^7+x^6+x^5+x^4+x^3+x^2+x$ & 1 \\
    Livermore-11 & $\frac{x^2x^3}{x+y}$ & 2 \\
    Livermore-12 & $\frac{x^5}{y^3}$ & 2 \\
    Livermore-14 & $x^3+x^2+x+\sin(x)+\sin(x^2)$ & 1 \\
    Livermore-15 & $x^{\frac{1}{5}}$ & 1 \\
    Livermore-16 & $x^{\frac{2}{5}}$ & 1 \\
    Livermore-17 & $4\sin(x)\cos(y)$ & 2 \\
    Livermore-18 & $\sin(x^2)\cos(x)-5$ & 1 \\
    Livermore-19 & $x^5+x^4+x^2+x$ & 1 \\
    Livermore-20 & $\exp(-x^2)$& 1 \\
    Livermore-21 & $x^8+x^7+x^6+x^5+x^4+x^3+x^2+x$ & 1 \\
    Livermore-22 & $\exp(-0.5x^2)$ & 1 \\
     \hline
    Koza2 & $x^5-2x^3+x$ & 1 \\
    Koza3 & $x^6-2x^4+x^2$ & 1 \\
     \hline
    Keijzer3 & $0.3x\sin(2\pi x)$ & 1  \\
    Keijzer4 & $x^3\exp(-x)\cos(x)\sin(x)(\sin(x^2)\cos(x)-1)$ & 1  \\
    Keijzer6 & $\frac{x(x+1)}{2}$ & 1 \\
    Keijzer7 & $\ln(x)$  & 1 \\
    Keijzer9 & $\ln(x+\sqrt{x^2+1})$ & 1  \\
    Keijzer11 & $xy+\sin{((x-1)(y-1))}$ & 2  \\
    Keijzer14 & $\frac{8}{2+x^2+y^2}$ & 2 \\
    Keijzer15 & $\frac{x^3}{5}+\frac{y^2}{2}-y-x$ & 2 \\
\end{longtable}
\end{footnotesize}

\begin{table}[]
    \centering
    % \footnotesize
    \caption{AI Feynman equations.}
    \label{AIFyenman}
    \begin{tabular}{ccc}
    \hline
    \textbf{Feynman Eq.} & \textbf{Equation} & \textbf{Variable}\\
    \hline
    I.6.20a  & $f=e^{-\theta^2/2}/\sqrt{2\pi}$ & 1 \\
        I.6.20   & $e^{-\frac{\theta^2}{2\sigma^2}}/\sqrt{2\pi \sigma^2}$ & 2  \\
        I.12.1   & $F=\mu N_n$  & 2 \\
        I.12.5   & $F=q_2E_f$  & 2 \\
        I.25.13  & $V_{e}=\frac{q}{c}$  & 2 \\
        I.29.4   & $k=\frac{\omega}{c}$  & 2 \\
        I.34.27  & $E=\hbar\omega$   & 2 \\
        I.39.10  & $E=\frac{3}{2}p_{F}V$  & 2  \\
        II.3.24  & $F_{E}=\frac{p}{4\pi r^{2}}$  & 2 \\
        II.8.31  & $E_{\mathrm{den}}=\frac{\epsilon E_{f}^{2}}{2}$  &  2 \\
        II.11.28 & $\theta=1+\frac{n\alpha}{1-n\alpha/3}$  &  2\\
        II.27.18 & $E_{\mathrm{den}}=\epsilon E_f^2$  & 2 \\
        II.38.14 & $\mu_S=\frac{\gamma}{2(1+\sigma)}$  & 2 \\
        III.12.43& $L=n\hbar$  & 2 \\\hline
    \end{tabular}
\end{table}

\section{Implement details}
\label{implement_details}
\setcounter{table}{0} 
\renewcommand{\thetable}{B\arabic{table}}
The details of our experimental setting is listed in Table \ref{experiment_setting}. We use the SetTransformer \cite{lee2019set} as our encoder for the output of the symbolic regression algorithm should be irrelevant with the input order. It requires the encoder has permutation invariance characteristic. SetTransformer satisfies the requirements as well as the ability to train with large scale training dataset.

\begin{table}[]
    \centering
    \caption{Experiment settings of our model.}
    \label{experiment_setting}
    \begin{tabular}{l l c}
        \hline
        \multicolumn{2}{c}{\textbf{Hyperparameter}} & \textbf{Value} \\\hline
        \multirow{4}{*}{Encoder} & head number & 12 \\
                                 & layer number & 4 \\
                                 & hidden dimension & 384 \\
                                 & dropout rate & 0.1 \\ \hline
        \multirow{4}{*}{Decoder} & hidden dimension & 384 \\
        & head number & 12 \\
        & layer number & 4 \\
        & maximum expression length & 50 \\ \hline
    \multirow{2}{*}{Batch size} & uni-variable & 256 \\
    & bi-variable & 128 \\ \hline
    \multirow{2}{*}{Training data} & uni-variable & 5,000,000 \\
    & bi-variable & 10,000,000 \\ \hline
    Beam search & beam size & 32 \\\hline
    \multicolumn{2}{l}{Maximum training epoch}  & 300 \\ \hline
    \multicolumn{2}{l}{Label smoothing} & 0.1 \\\hline
    \multicolumn{2}{l}{$\eta$ (training )} & 0.1 \\\hline
    \multicolumn{2}{l}{$\lambda$} & 0.1 \\\hline
    \end{tabular}
\end{table}

We run our models in cloud devices, which contains 4 A100 GPUs. The training is early stopped if the train loss is converged.

\section{Baseline details}
\label{baseline_details}
\textbf{Large-scale pre-trained models.} We compare the performance of our method with four strong symbolic regression baselines that belong to large-scale pre-trained model:
\begin{itemize}
    \item \textbf{NeSymReS \cite{DBLP:conf/icml/BiggioBNLP21}.} A standard Transformer-based pre-trained model, it feeds the data points into the networks and uses the pre-order traversal as the expression label to learn the map relationship between data points and skeletons. BFGS is utilized to optimize the constant value in the skeleton.
    \item \textbf{E2E \cite{kamienny2022end}.} Based on the NeSymReS, E2E design a strategy to encode constant, the constant is learned during the training process. Therefore, E2E provides a better constant initial guess for constant optimization.
    \item \textbf{T-JST \cite{li2022transformer}.} To tackle with the ill-posed problem of NeSymReS, T-JST apply contrastive loss to the training to enclose the distance between the data features derived from same skeleton, and enlarge the distance between two different skeletons.
    \item \textbf{DeepSymNet \cite{10327762}.} A recently publicated pre-trained model that encode expressions by a designed encoding strategy. DeepSymNet uses SymNet (a special form of graph) to represent expressions.
\end{itemize}

\textbf{End-to-end models.} GP-based symbolic regression, deep reinforcement learning-based methods,and others are belong to the end-to-end models. We compare with the methods included in SRBench. For a detailed description of the method and its implementation, please refer to SRBench \cite{DBLP:journals/corr/abs-2107-14351}.

\section{Complementary experimental results}
\label{Complementary_experimental_results}
\setcounter{figure}{0} 
\renewcommand{\thefigure}{C\arabic{figure}}

\begin{figure}[htbp]
    \centering
    \begin{subfigure}[b]{0.3\textwidth}
        \includegraphics[width=\textwidth]{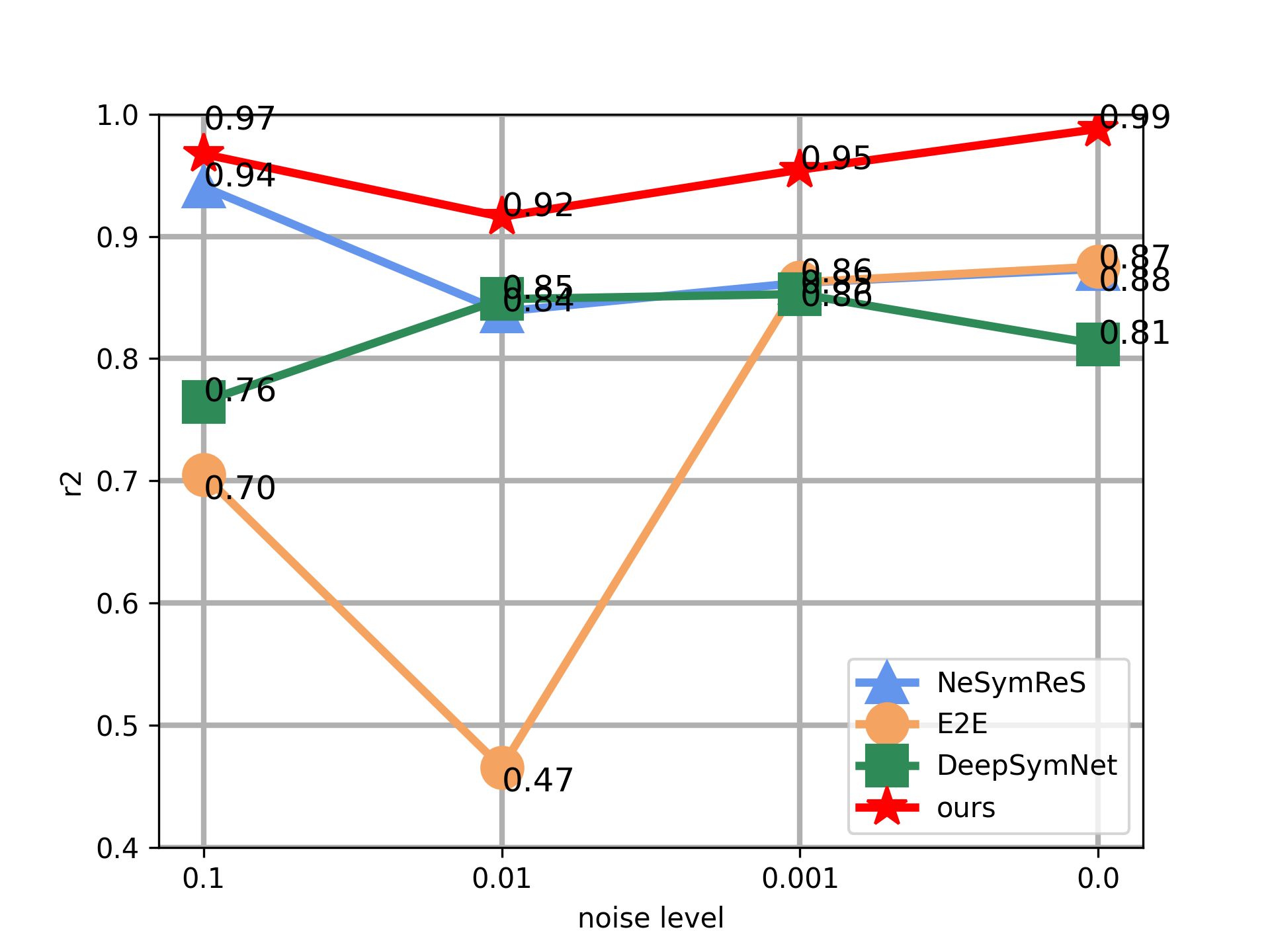}
        \caption{Keijizer}
        \label{Keijzer}
    \end{subfigure}
    \hfill
    \begin{subfigure}[b]{0.3\textwidth}
        \includegraphics[width=\textwidth]{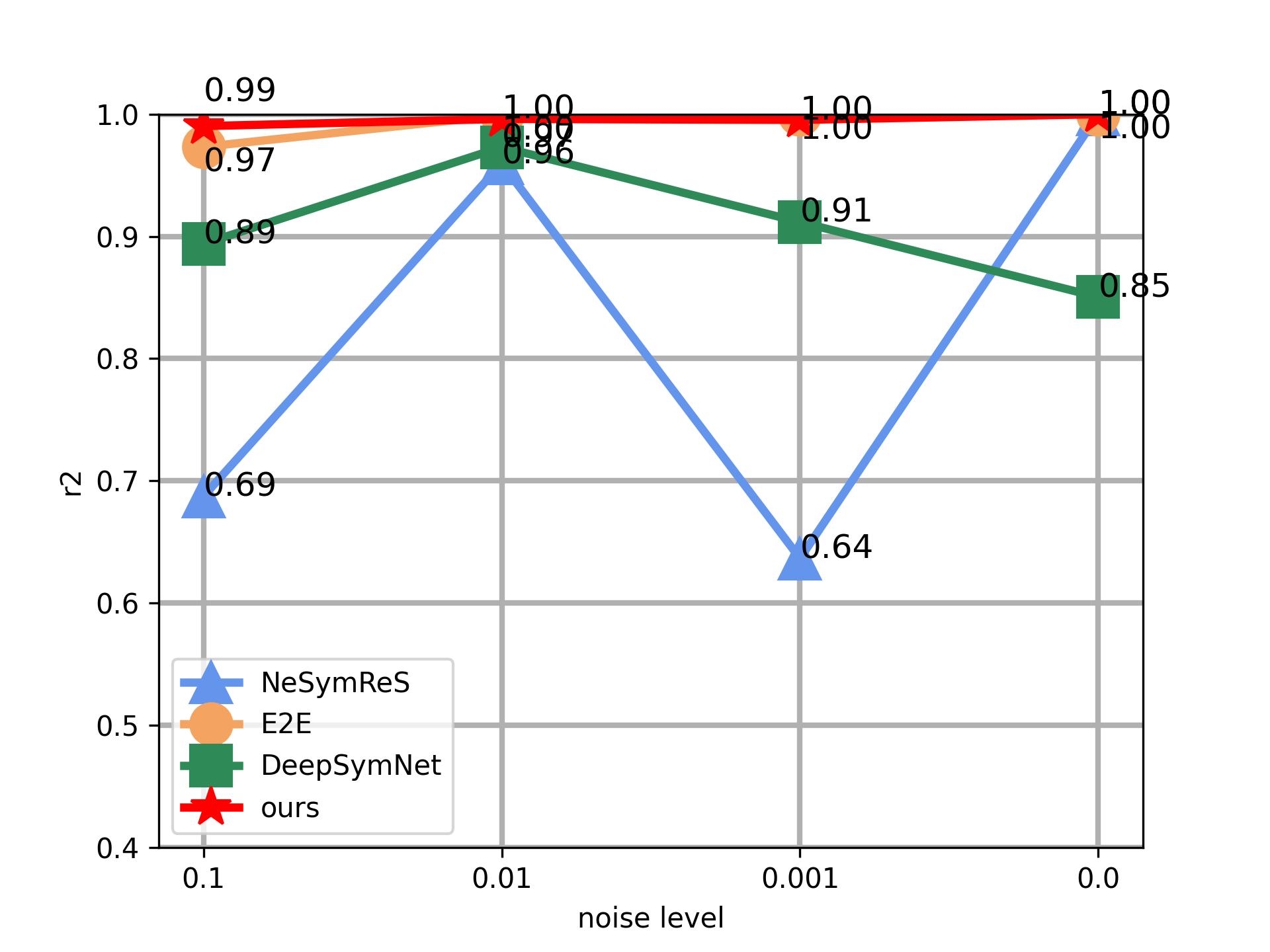}
        \caption{Koza}
        \label{Koza}
    \end{subfigure}
    \hfill
    \begin{subfigure}[b]{0.3\textwidth}
        \includegraphics[width=\textwidth]{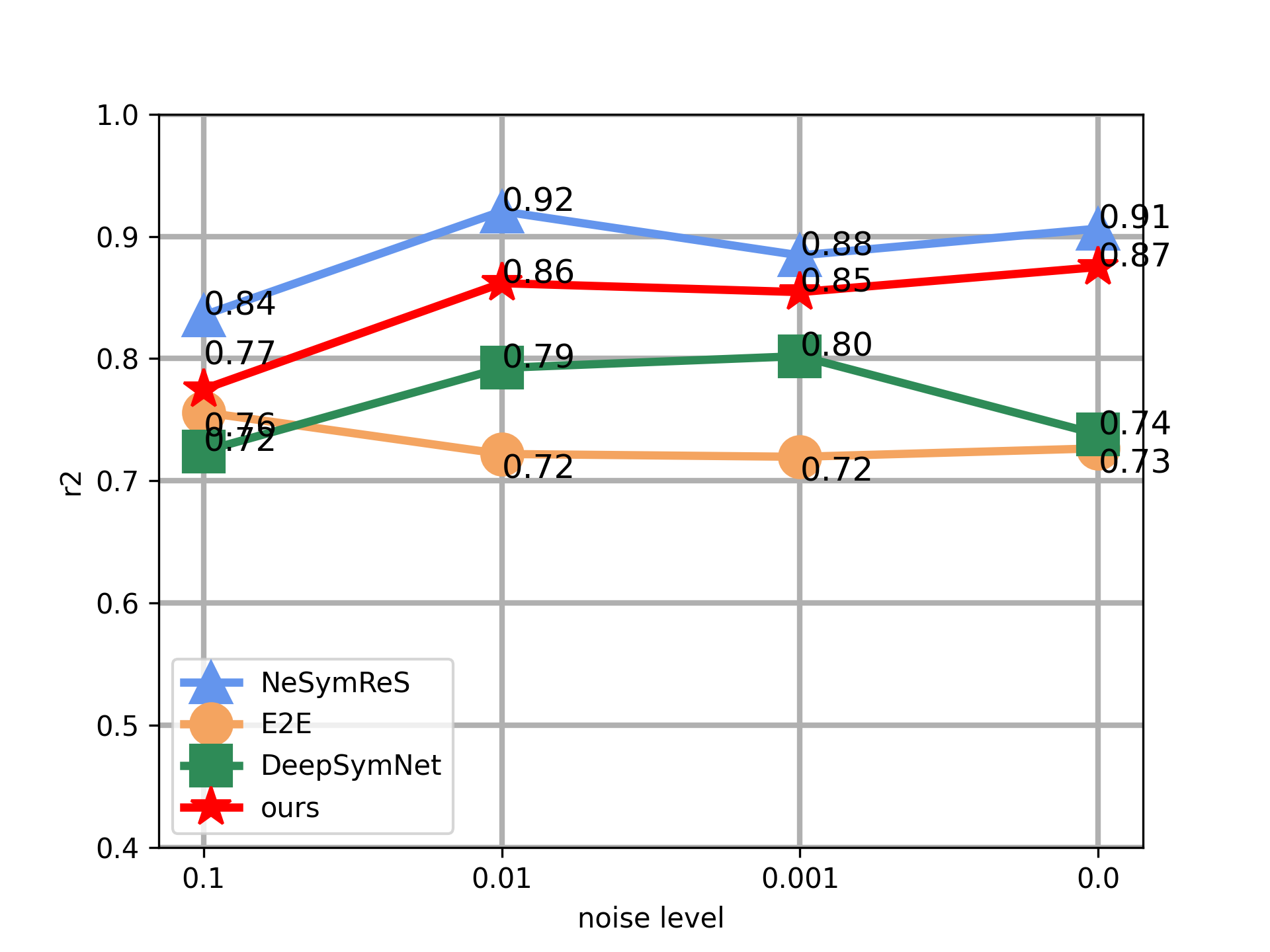}
        \caption{Constant}
        \label{Constant}
    \end{subfigure}
    \\
    \begin{subfigure}[b]{0.3\textwidth}
        \includegraphics[width=\textwidth]{./figs/Livermore.png}
        \caption{Livermore}
        \label{Livermore}
    \end{subfigure}
    \hfill
    \begin{subfigure}[b]{0.3\textwidth}
        \includegraphics[width=\textwidth]{./figs/Nguyen.png}
        \caption{Nguyen}
        \label{Nguyen}
    \end{subfigure}
    \hfill
    \begin{subfigure}[b]{0.3\textwidth}
        \includegraphics[width=\textwidth]{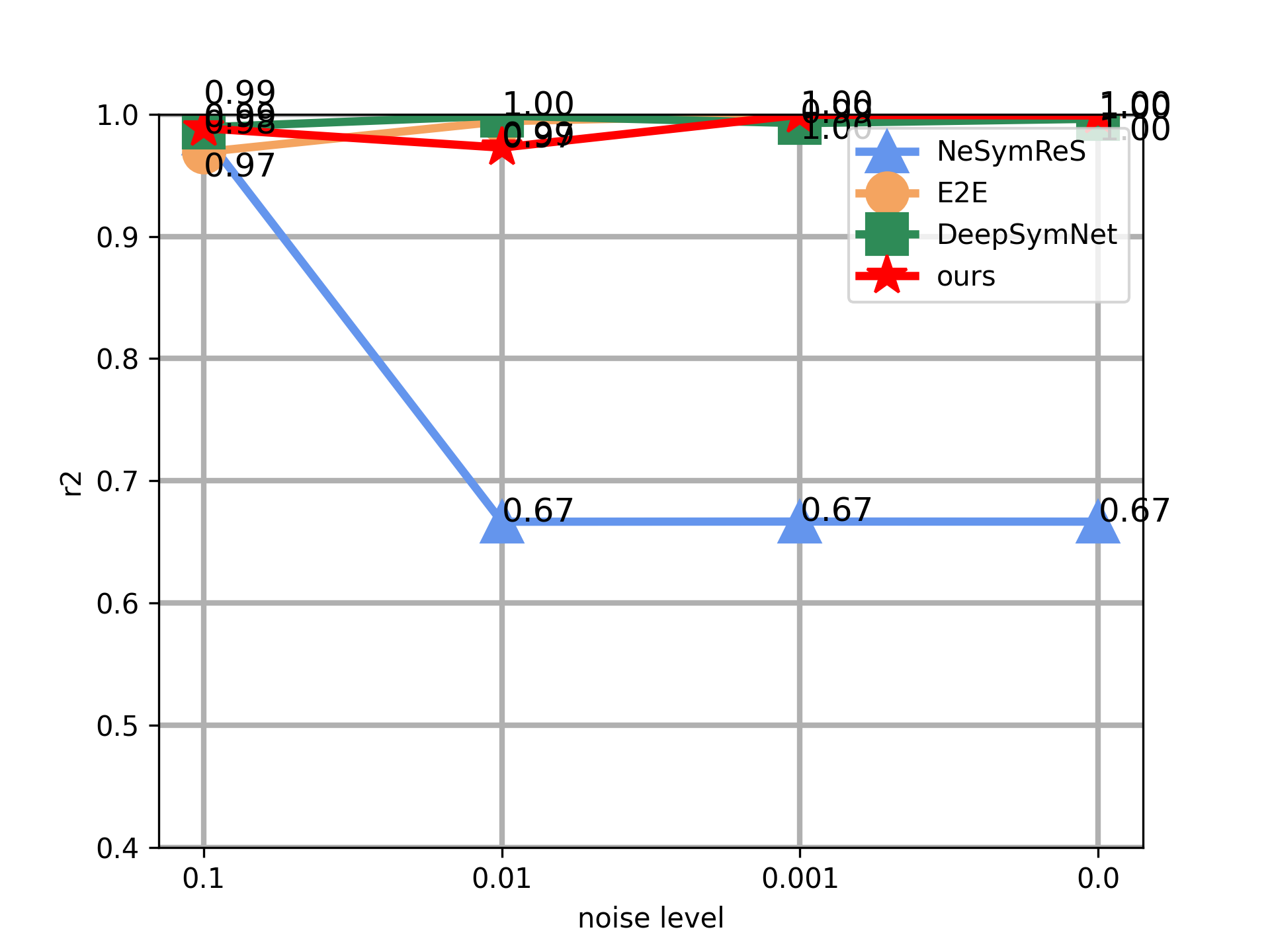}
        \caption{R}
        \label{R}
    \end{subfigure}

    \caption{Mean $R^2$ of different methods under different noise levels. The noise level is ordered as [0.1, 0.01, 0.001, 0.0]. The red, blue, green, and brown line represents our model, NeSymReS, DeepSymNet, and E2E, respectively.}
    \label{complement_figures}
\end{figure}

The results in the Figure \ref{complement_figures} is the complement to the section \ref{compared_with_pre_trained}. The $x$-axis is the noise level, ordered by descendent values [0.1, 0.01, 0.001, 0.0]. The $y$-axis is the mean $R^2$ of each dataset. Our model achieves superior performances against others in many datasets in different levels of noise. It illustrates that our model could handle with noise data.

\end{document}